\documentclass[sigconf, pbalance]{acmart}
\usepackage{multirow}

\usepackage{subcaption}
\usepackage{pifont}
\usepackage{colortbl}
\usepackage{placeins}

\newcommand{\cmark}{\textcolor{green!60!black}{\ding{51}}}
\newcommand{\xmark}{\textcolor{red!75!black}{\ding{55}}}

\definecolor{RankFirst}{HTML}{D5E8D4}   
\definecolor{RankSecond}{HTML}{FFF2CC}  
\definecolor{RankThird}{HTML}{DAE8FC}   

\AtBeginDocument{%
  }

\copyrightyear{2026}
\acmYear{2026}
\setcopyright{cc}
\setcctype{by}
\acmConference[MM '26]{Proceedings of the 34th ACM International Conference on Multimedia}{November 10--14, 2026}{Rio de Janeiro, Brazil}
\acmBooktitle{Proceedings of the 34th ACM International Conference on Multimedia (MM '26), November 10--14, 2026, Rio de Janeiro, Brazil}
\acmISBN{979-8-4007-2213-4/2026/11}
\acmDOI{10.1145/3767308.3834980}

\begin{document}

\title{Diffusion Image Editing via Asynchronous Token Decoding}

\author{Yang Shi}
\authornote{Yang Shi, Liangsi Lu, and Minzhe Guo contributed equally to this work.}
\affiliation{%
  \institution{Guangdong University of Technology}
  \city{Guangzhou}
  \country{China}
}
\email{sudo.shiyang@gmail.com}
\orcid{0009-0009-3928-7495}

\author{Liangsi Lu}
\authornotemark[1]
\authornote{Corresponding author.}
\affiliation{%
  \institution{Guangdong University of Technology}
  \city{Guangzhou}
  \country{China}
}
\email{lu.liangsi.cn@gmail.com}
\orcid{0009-0006-2839-3901}

\author{Minzhe Guo}
\authornotemark[1]
\affiliation{%
  \institution{Guangdong University of Technology}
  \city{Guangzhou}
  \country{China}
}
\email{guominzhe@mails.gdut.edu.cn}
\orcid{0009-0001-2511-4763}

\author{Yifeng Xie}
\affiliation{%
  \institution{Hong Kong Baptist University}
  \city{Hong Kong}
  \country{China}
}
\email{evfxie@gmail.com}
\orcid{0009-0008-8333-9419}

\author{Yanhui Chen}
\affiliation{%
  \institution{Guangdong University of Technology}
  \city{Guangzhou}
  \country{China}
}
\email{chenyanhui91@mails.gdut.edu.cn}
\orcid{0009-0004-8595-1274}

\author{Jingchao Wang}
\affiliation{%
  \institution{Peking University}
  \city{Beijing}
  \country{China}
}
\email{ethanwangjc@163.com}
\orcid{0000-0002-0099-539X}

\author{Xuhang Chen}
\affiliation{%
  \institution{Huizhou University}
  \city{Huizhou}
  \country{China}
}
\email{xuhangc@hzu.edu.cn}
\orcid{0000-0001-6000-3914}

\renewcommand{\shortauthors}{Shi et al.}

\begin{abstract}
Text-guided diffusion image editing aims to modify semantic attributes of an image while preserving its identity, layout, and background. However, na\"ively switching the text condition during sampling often causes global drift, as denoising dynamics propagate changes across tokens and can disrupt unedited regions. To address this issue, we propose \textbf{A}synchronous \textbf{T}oken \textbf{D}ecoding \textbf{Edit} (ATDEdit), an inference-time framework that views each sampler step as a parallel update of a globally coupled token matrix and enables token-indexed condition switching with differentiated update policies. Instead of applying synchronous target-conditioned updates to all tokens, ATDEdit estimates editable locations using token-wise conditional surprisal and applies target-conditioned corrections to the selected token set. It supplies source key/value memory at keep-token positions and projects selected keep-token latent rows back to their source values; these operations promote background preservation but do not constitute a pixel-level invariance guarantee. This approach combines local editing and background preservation without external or user-provided spatial masks and without model fine-tuning. On PIE-Bench, ATDEdit achieves the strongest reported preservation metrics, including 27.44~dB PSNR and 0.055 LPIPS, while retaining competitive semantic alignment.
\end{abstract}

\ccsdesc[500]{Computing methodologies~Computer vision problems}
\ccsdesc[300]{Computing methodologies~Reconstruction}

\keywords{Text-Guided Image Editing, Diffusion Models, Diffusion Transformers, Asynchronous Token Decoding, Inference-Time Editing}
\begin{teaserfigure}
  \includegraphics[width=\textwidth]{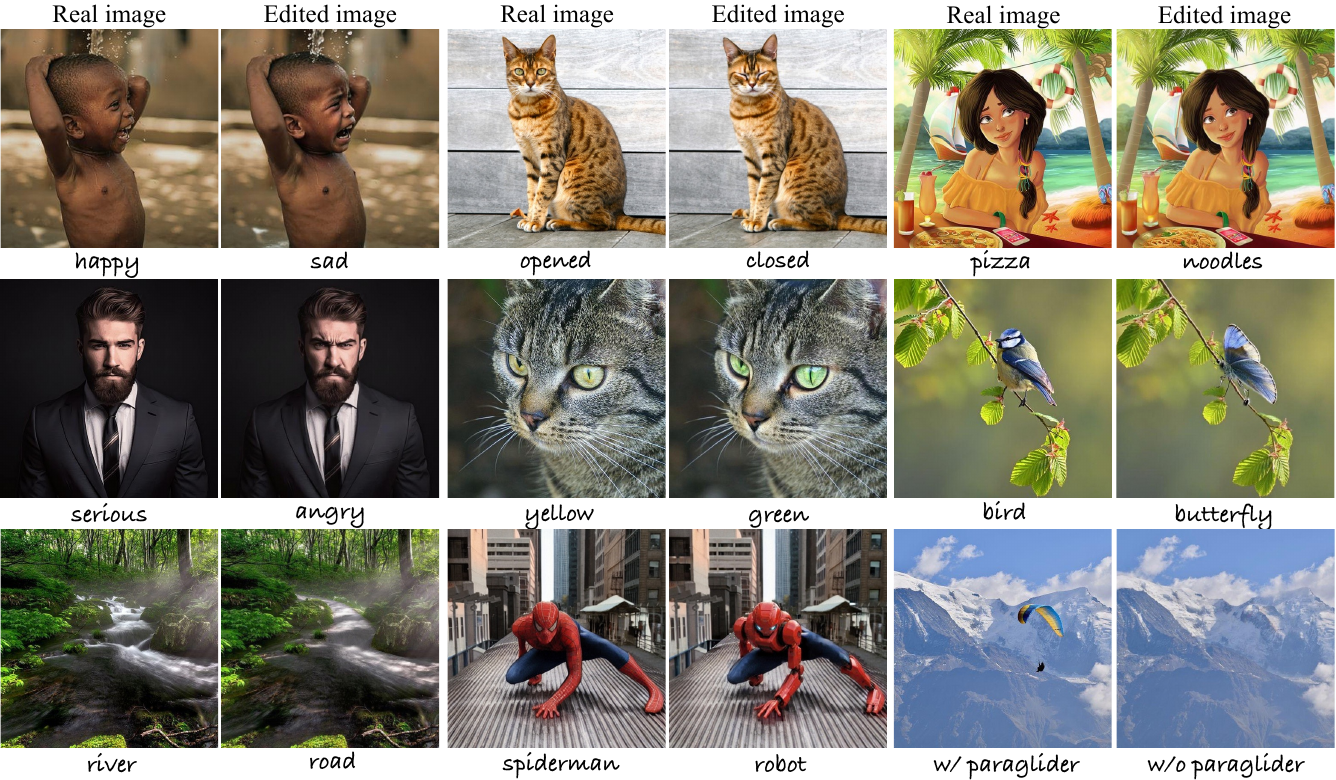}
  \caption{\textbf{ATDEdit}. These examples demonstrate our inference-time, training-free framework operating on Diffusion Transformers. ATDEdit mitigates the global drift of synchronous decoding through token-wise condition switching driven by Token-wise Conditional Surprisal. Source-memory replacement and hard projection promote preservation of selected keep-token regions while editable tokens receive target-conditioned updates. Labels indicate the desired semantic change.}
  \Description{A grid of source and edited images illustrating several local semantic edits. The intended object or attribute changes while the surrounding layout, lighting, and background remain visually stable.}
  \label{fig:teaser}
\end{teaserfigure}

\maketitle

\section{Introduction}

\begin{figure}[!t]
    \centering
    \includegraphics[width=1\linewidth]{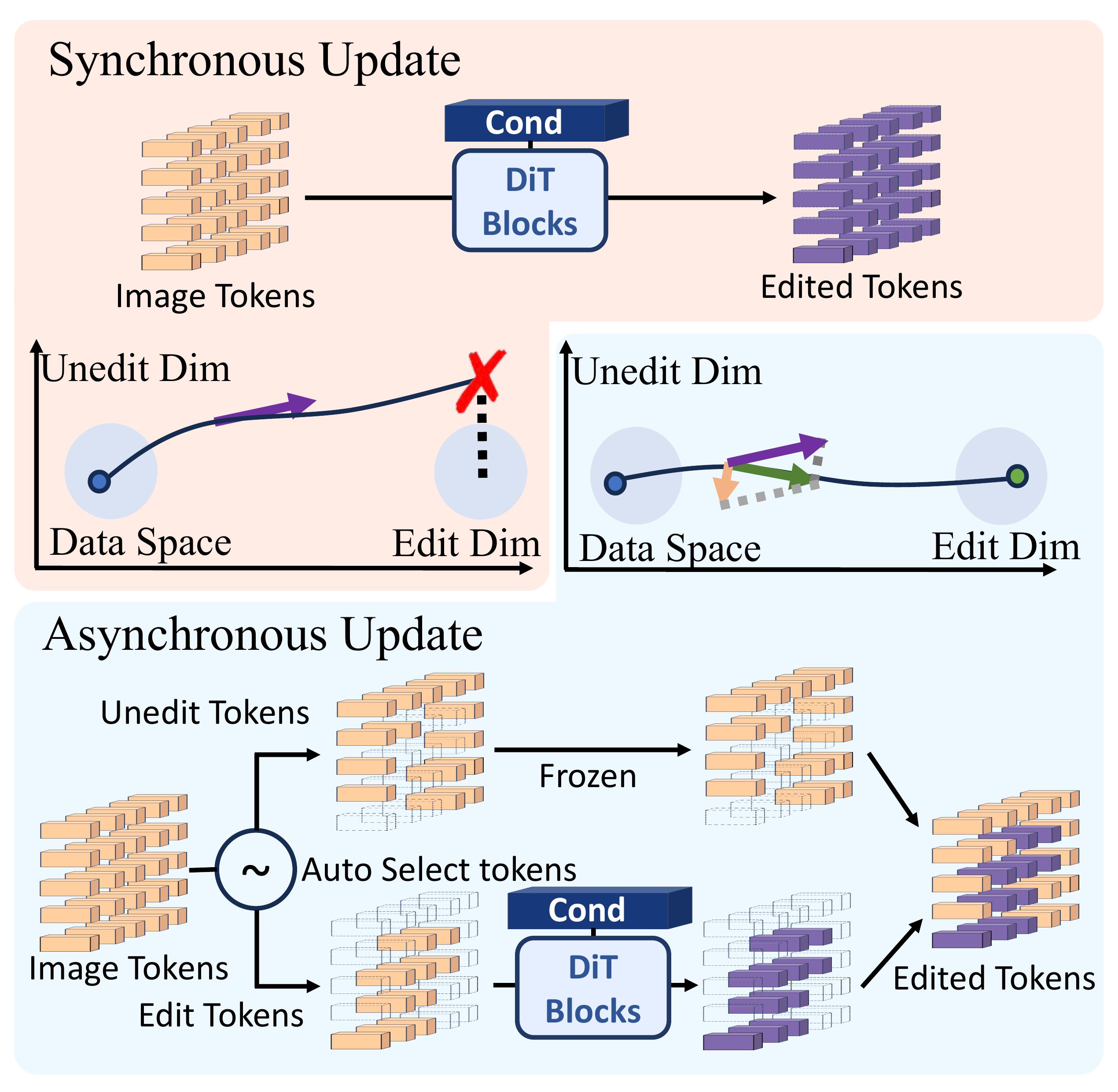}
    \caption{\textbf{Comparison between Synchronous and Asynchronous Token Decoding.} 
  \textbf{Top:} Conventional synchronous editing updates all latent tokens uniformly under the target condition. This uniform update causes the denoising trajectory to diverge globally (marked by \xmark), leading to unwanted background drift. 
  \textbf{Bottom:} Our framework reformulates editing as an asynchronous process. It dynamically utilizes Token-wise Conditional Surprisal to identify editable regions. Selected keep-token latent rows are projected back to their source values after each step, while editable tokens receive the target-conditioned correction.}
    \Description{A two-part conceptual diagram. The top shows synchronous decoding updating every image token and causing unwanted background drift. The bottom shows ATDEdit updating only high-surprisal editable tokens while source-projecting selected keep-token rows.}
    \label{fig:async_concept}
\end{figure}

Text-guided image editing has become a crucial tool across a range of domains~\cite{meng2021sdedit,brooks2023instructpix2pix}. Its core mission is to empower users to modify specific semantic attributes of an image, such as altering the style of clothing, changing the time of day, or transforming the breed of an animal. Driven by the demand for high-fidelity results and fine-grained control, research in this area is rapidly advancing~\cite{cao2023masactrl,zhu2025kv}.

Recently, the field has witnessed a paradigm shift with the advent of Diffusion Transformers (DiTs)~\cite{peebles2023scalable}, which have largely supplanted U-Nets as the architecture of choice for high-quality synthesis. 
Building upon DiTs, a compelling line of work such as FlowEdit~\cite{kulikov2025flowedit} has further refined the process by moving beyond stochastic noise loops. These approaches reinterpret diffusion as a deterministic transport via Ordinary Differential Equations (ODEs)~\cite{song2020score} or Rectified Flows~\cite{liu2022rectified}, effectively leveraging the transformer's inherent parallelism to map source latents directly to target states. This theoretically offers a precise trajectory for semantic transformation.

However, despite these advancements, a fundamental limitation persists: many trajectory-based editors treat the decoding trajectory as a monolithic, synchronous transformation. By updating all latent tokens uniformly under the new prompt, they do not exploit the token-indexed latent representation of DiTs; importantly, these tokens are still globally coupled through self-attention. Thus, modifying the prompt (e.g., from ``dog'' to ``cat'') and imposing a global update can trigger unwanted changes outside the intended region. Conversely, methods relying on pre-computed static masks (e.g., DiffEdit)~\cite{couairon2022diffedit} decouple the ``where'' from the ``what'' of generation. Yet, imposing rigid spatial constraints onto a dynamic process often fails to capture the scene's subtle, evolving semantics~\cite{yan2026less, yan2025entropy}.

To address these limitations, we propose to shift the paradigm from imposing static external masks to enabling dynamic internal model introspection.
In DiTs~\cite{peebles2023scalable}, we formulate diffusion editing as an asynchronous token-indexed process: all tokens remain coupled through the transformer, but different token rows may follow different condition-switching and state-update rules. This token-indexed representation provides the structural basis for trajectory-level asynchrony without assuming statistical independence between tokens.
Guided by this, we introduce \textbf{A}synchronous \textbf{T}oken \textbf{D}ecoding \textbf{Edit} (ATDEdit). Instead of imposing external constraints, ATDEdit uses the model's token-wise counterfactual response to ask which locations are most sensitive to the new prompt. We quantify this using Token-wise Conditional Surprisal, a token-wise conditional sensitivity score measuring the magnitude of prediction change under a condition switch. This reformulates editing as a negotiation: high-surprisal tokens are permitted to receive target-conditioned updates, while selected keep-token latent states are reset to their source values and their key/value memory is supplied from the source branch. These operations reduce unwanted changes without claiming that the underlying transformer dynamics are decoupled.
Figure~\ref{fig:teaser} visually demonstrates the efficacy of ATDEdit.

Figure~\ref{fig:async_concept} contrasts this token-indexed update with conventional synchronous decoding. We provide a scoped theoretical analysis. A deterministic sampler is a Markov recursion that updates the full token matrix in parallel, but this does not imply token independence. For common-noise coupling, the scalar difference has no larger variance than independent-noise evaluation when the paired source/target evaluations have nonnegative covariance, and strictly smaller variance when the covariance is positive. For hard projection, keep-token latent states are exactly reset to the source state at each step in exact arithmetic; this is a token-state guarantee rather than a pixel-level invariance claim.

In summary, our key contributions are as follows: 

\begin{itemize}
    \item We propose ATDEdit, a framework that reformulates diffusion editing as asynchronous token decoding. Central to this is Token-wise Conditional Surprisal, a token-wise conditional sensitivity score that dynamically discovers editable regions online by quantifying each token's response to a condition switch. 
    \item We introduce an asynchronous update mechanism integrating an Attention Memory Boundary Condition and Common-noise Coupling. Hard projection exactly resets selected keep-token latent states, while common noise aligns stochastic inputs and can reduce scalar estimator variance under a nonnegative-covariance condition.
    \item We clarify the scope of the method's formal properties: sampler updates are parallel but globally coupled, top-quantile selection optimizes captured surprisal only under a fixed cardinality budget, and the zero-drift result applies to projected keep-token latent states.
    \item We conduct extensive experiments on PIE-Bench, where ATDEdit obtains the strongest preservation metrics in the reported comparison while retaining competitive semantic alignment.
\end{itemize}

\section{Related Work}

\subsection{Diffusion and Flow Backbones}
Diffusion models~\cite{ho2020denoising,song2020score, yan2026pixel} and latent diffusion models~\cite{rombach2022high} enable high-quality generation, while recent Diffusion Transformers and flow-based formulations further improve scalability and sampling efficiency~\cite{peebles2023scalable,liu2022flow}. More broadly, recent world foundation models such as Orca formulate multimodal understanding and generation through a unified latent state-transition space~\cite{wang2026orca}. However, when the condition is changed within one reverse trajectory, existing samplers still update all tokens synchronously, which can propagate undesired changes beyond the edited region~\cite{zhu2025kv}. In contrast, ATDEdit keeps the backbone unchanged and modifies only the inference dynamics through token-asynchronous condition switching.

\subsection{Text Guided Image Editing}
Text-guided image editing methods include noise-based approaches such as SDEdit~\cite{meng2021sdedit}, attention-control methods~\cite{hertz2022prompt,tumanyan2023plug,cao2023masactrl}, inversion-based methods~\cite{mokady2023null,kawar2023imagic}, and instruction-following models~\cite{brooks2023instructpix2pix}. Closely related composed retrieval work also combines visual content with modification text, with recent methods studying attribute-aware composition, robustness to noisy correspondences, and directional calibration across image/video retrieval~\cite{COMBINER,ConeSep,ReTrack}. Recent token-level controls~\cite{wang2024tokencompose,kamenetsky2025saedit} provide finer spatial control, but they rely on additional training and therefore differ from a training-free inference-time setting. A persistent difficulty is to improve instruction alignment without sacrificing background preservation under synchronous conditioning~\cite{cao2023masactrl, lu2026chordedit, lu2026semantic}. In contrast, ATDEdit is training free and applies target-conditioned updates only to tokens selected online during sampling.

\subsection{Mask Discovery and Attention Based Preservation}
Mask-based approaches such as DiffEdit~\cite{couairon2022diffedit} and memory-based DiT editors such as KVEdit~\cite{zhu2025kv} constrain generation through discovered masks or attention reuse. PartEdit~\cite{cvejic2025partedit} further targets object-part editing by optimizing part-specific textual tokens to obtain localization masks during inference. Yet most of these methods depend on offline masks, external priors, or static heuristics, and they do not adapt the protected region to the time-varying reverse dynamics~\cite{lu2026semantic}. In contrast, ATDEdit discovers editable tokens online and combines this discovery with an attention memory boundary and a state-space projection.

\section{Method}
\label{sec:method}

\begin{figure*}
    \centering
    \includegraphics[width=1\linewidth]{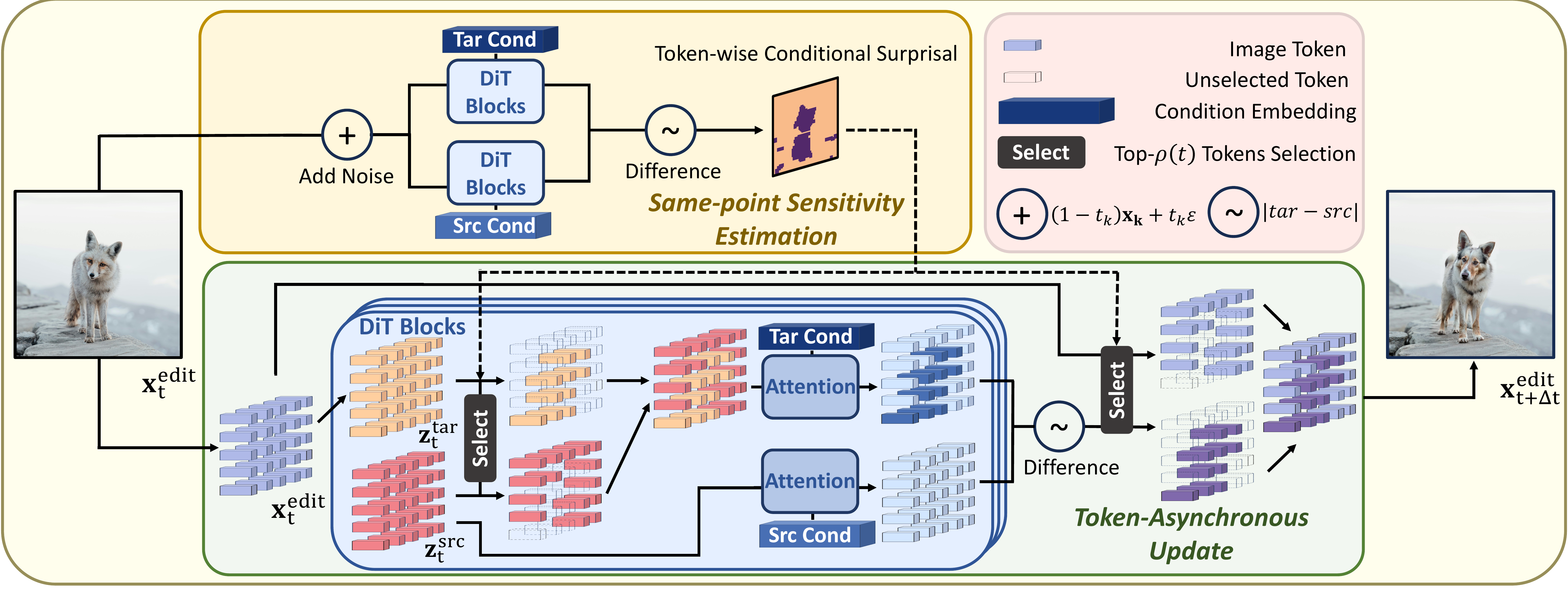}
    \caption{\textbf{Overview of ATDEdit for transforming a fox into a dog.} \textbf{(Top)} Same-state source/target evaluations produce Token-wise Conditional Surprisal and an editable-token mask. \textbf{(Bottom)} Target-conditioned corrections are applied to editable tokens, while source key/value memory and hard projection constrain selected keep-token rows.}
    \Description{A method pipeline for changing a fox into a dog. The upper stage computes token-wise conditional surprisal and selects editable tokens. The lower stage applies target-conditioned updates to selected tokens while source key-value memory and hard projection preserve the background.}
    \label{fig:method}
\end{figure*}

In DiT or flow samplers, switching the condition from $c_{\mathrm{src}}$ to $c_{\mathrm{tar}}$ within one sampling trajectory is equivalent to asking all tokens to synchronously rewrite their denoising explanations.
However, a local edit may require only a subset of tokens to change, and forcing a synchronous rewrite can cause global drift on tokens that should remain stable~\cite{hertz2022prompt}.
A concrete prediction is that the same-state counterfactual response is strongly token-nonuniform and varies over time, defining a time-dependent editable set $\mathcal{E}(t)$.
However, this formulation faces two challenges. First, self-attention couples tokens globally; even if we never explicitly update keep tokens in state space, they can still drift through message passing unless an internal boundary condition is imposed.
Second, counterfactual differences estimated under mismatched randomness have high variance, destabilizing both $\mathcal{E}(t)$ selection and target-conditioned updates~\cite{yang1991using}.
To address these limitations, as illustrated in Figure~\ref{fig:method}, our ATDEdit framework reformulates the editing process into two parallel stages: sensitivity estimation and asynchronous updating.

\subsection{Preliminaries and Notation}

We use a token-indexed view of diffusion sampling. A deterministic sampler maps the current full token matrix to the next full token matrix, so token rows are updated simultaneously at each scheduler step. Self-attention nevertheless couples those rows; our ``asynchronous'' formulation means that token indices can receive different condition-switching and state-update rules, not that their dynamics are independent. This view is conceptually related to parallel token decoding in masked generative transformers for text-to-image generation~\cite{chang2023muse}, while our formulation concerns continuous latent states in diffusion/flow samplers.

We therefore describe diffusion sampling in DiTs as parallel token-state decoding along sampling time.
Let $\mathbf{z}_t \in \mathbb{R}^{S \times d}$ denote the latent token sequence at continuous time $t\in[0,1]$, where $S$ is the number of tokens and $d$ is the token feature dimension; larger $t$ corresponds to noisier latents.
For any matrix $\mathbf{A}\in\mathbb{R}^{S\times d}$ and any subset $\mathcal{J}\subseteq\{1,\dots,S\}$, we write $\mathbf{A}_{\mathcal{J}}\in\mathbb{R}^{|\mathcal{J}|\times d}$ for the row-submatrix restricted to indices in $\mathcal{J}$.
For a single index $i\in\{1,\dots,S\}$, $\mathbf{A}^{(i)}\in\mathbb{R}^{d}$ denotes the $i$-th token row.
We follow common diffusion notation and use bold lowercase for token sequences, which are $S\times d$ matrices; capital letters are reserved for attention matrices.
Let $c_{\mathrm{src}}$ and $c_{\mathrm{tar}}$ denote the source and target conditions, and write $c$ for a generic condition.
We use superscripts $\mathrm{src}/\mathrm{tar}$ to indicate the source/target branches, and subscripts $t$ or $t_j$ to denote scheduler time.

A diffusion transformer with parameters $\theta$ can be abstractly written as a one-step decoder
\begin{equation}
    \hat{\mathbf{z}}_{t_{j+1}} = \mathcal{D}_\theta(\mathbf{z}_{t_j}, t_j, c),
\end{equation}
where $\{t_j\}_{j=1}^{N+1}$ is a decreasing scheduler sequence with $t_1 > t_2 > \cdots > t_{N+1}$ and $N$ is the number of reverse inference steps. We define the signed step size
\begin{equation}
    \Delta t_j \triangleq t_{j+1} - t_j,
\end{equation}
so $\Delta t_j < 0$ for a decreasing schedule. We use a rectified-flow parameterization~\cite{liu2022flow}, in which the model predicts a velocity field
\[
v_\theta(\mathbf{z}_t, t, c) \in \mathbb{R}^{S \times d}.
\]

Under this parameterization, an estimate of the clean latent tokens $\mathbf{x}_0$ is recovered as
\begin{equation}
    \hat{\mathbf{x}}_0(\mathbf{z}_t,t,c) = \mathbf{z}_t - t\, v_\theta(\mathbf{z}_t,t,c),
    \label{eq:x0_from_v}
\end{equation}
which is the form used throughout this section.

Let $\mathbf{x}^{\mathrm{src}}_{t} \in \mathbb{R}^{S \times d}$ be the source $x_0$-space latent tokens (encoded from the source image). For notational alignment with the scheduler, we keep the subscript $t$ (or $t_j$), although $\mathbf{x}^{\mathrm{src}}_{t}$ is constant across $t$.
We maintain an editable $x_0$-space state $\mathbf{x}^{\mathrm{edit}}_{t} \in \mathbb{R}^{S \times d}$, and use $\mathbf{x}^{\mathrm{edit}}_{t_j}$ to denote the state associated with scheduler time $t_j$.
We initialize $\mathbf{x}^{\mathrm{edit}}_{t_1} \leftarrow \mathbf{x}^{\mathrm{src}}_{t_1}$.

We execute the reverse process on the decreasing scheduler sequence $\{t_j\}_{j=1}^{N+1}$ defined above.

We use $\mathbf{0},\mathbf{1}\in\mathbb{R}^S$ to denote the all-zeros and all-ones vectors.
For token-wise masks, we use broadcast over the feature dimension: for a mask vector
$\mathbf{m}\in\{0,1\}^S$ and a token matrix $\mathbf{A}\in\mathbb{R}^{S\times d}$,
$\mathbf{A}\odot \mathbf{m}$ denotes element-wise multiplication where $\mathbf{m}$ is broadcast to shape $S\times d$.
For an editable mask $\mathbf{m}(t)\in\{0,1\}^S$, define the editable and keep sets as
$\mathcal{E}(t)\triangleq\{i:m_i(t)=1\}$ and $\mathcal{K}(t)\triangleq\{i:m_i(t)=0\}$, with keep indicator $\mathbf{k}(t)\triangleq\mathbf{1}-\mathbf{m}(t)$.
We use the subscript ``keep'' to denote row selection by $\mathcal{K}(t)$: for a token matrix $\mathbf{A}\in\mathbb{R}^{S\times d}$, $\mathbf{A}_{\text{keep}}=\mathbf{A}_{\mathcal{K}(t)}$.

\subsection{Editing as a Constrained Token Dynamics}
\label{subsec:contract}

To formalize the part of background preservation that is directly controllable in latent state space, we define a keep-token constraint during the reverse process.
We express this constraint as a state-space contract and enforce it by projection.
Given a keep set $\mathcal{K}(t)\subseteq\{1,\dots,S\}$ at time $t$, define the affine constraint set
\begin{equation}
\mathcal{C}(t)\triangleq
\Big\{\mathbf{x}\in\mathbb{R}^{S\times d}:\ \mathbf{x}_{\mathcal{K}(t)}=(\mathbf{x}^{\mathrm{src}}_{t})_{\mathcal{K}(t)}\Big\}.
\end{equation}
We enforce this contract by Euclidean projection
\begin{equation}
\Pi_{\mathcal{C}(t)}(\mathbf{y})\triangleq\arg\min_{\mathbf{x}\in\mathcal{C}(t)}\|\mathbf{x}-\mathbf{y}\|_F,
\end{equation}
where $\|\cdot\|_F$ denotes the Frobenius norm on matrices, $\|\mathbf{A}\|_F \triangleq \sqrt{\sum_{i=1}^S\sum_{\ell=1}^d A_{i\ell}^2}$.
Separating the Frobenius objective over keep and editable rows shows that Eq.~\eqref{eq:hard_reset} is the Euclidean projection onto this constraint set. It therefore yields exact zero drift for the currently selected keep-token latent rows in the ideal arithmetic model. The remaining question is how to identify $\mathcal{E}(t)$ online and how to compute a stable counterfactual update on that set.

\subsection{Common-Noise Coupling and Parallelogram Sampling}\label{sec:coupling}
Reliable editing requires reducing noise mismatch when estimating the response associated with a condition switch, because stochastic variations can dominate comparisons based on independent noise.
We construct the counterfactual comparison between $c_{\mathrm{src}}$ and $c_{\mathrm{tar}}$ under aligned randomness using a common-noise coupling.

Let $\boldsymbol{\varepsilon} \in \mathbb{R}^{S \times d}$ denote Gaussian noise with i.i.d.\ standard normal entries, equivalently
$\operatorname{vec}(\boldsymbol{\varepsilon})\sim\mathcal{N}(\mathbf{0}_{Sd},I_{Sd})$.
At time $t$, we construct a source noisy latent using the rectified-flow linear interpolation:
\begin{equation}
    \mathbf{z}^{\mathrm{src}}_t = (1-t)\,\mathbf{x}^{\mathrm{src}}_{t} + t\,\boldsymbol{\varepsilon}.
    \label{eq:z_src}
\end{equation}

Rather than sampling an independent target trajectory, we couple the target noisy latent by a parallelogram construction~\cite{kulikov2025flowedit}:
\begin{equation}
    \mathbf{z}^{\mathrm{tar}}_t = \mathbf{z}^{\mathrm{src}}_t + \left(\mathbf{x}^{\mathrm{edit}}_{t} - \mathbf{x}^{\mathrm{src}}_{t}\right),
    \label{eq:z_tar}
\end{equation}
so both branches share the same noise realization $\boldsymbol{\varepsilon}$ while differing only by the current editable displacement
$\mathbf{x}^{\mathrm{edit}}_{t} - \mathbf{x}^{\mathrm{src}}_{t}$.
This coupling removes branch-wise randomness mismatch and can reduce estimator variance. For scalar target/source functionals $F$ and $G$, the coupled difference satisfies $\mathrm{Var}(F-G)=\mathrm{Var}(F)+\mathrm{Var}(G)-2\mathrm{Cov}(F,G)$; relative to independent noise, its variance is no larger when $\mathrm{Cov}(F,G)\ge 0$ and is strictly smaller when the covariance is positive. The difference can still reflect the condition switch, the current state displacement, and the applied memory boundary.

\subsection{Online Editable-Token Discovery via Token-wise Conditional Surprisal}
\label{sec:surprisal}

We characterize what the dynamic masking rule guarantees and what it does not. Conditional surprisal is exactly a scaled same-state velocity difference under Eq.~\eqref{eq:x0_from_v}. Under a fixed cardinality budget, selecting the largest scores maximizes only the sum of selected surprisal values; it is not a claim of global optimality for the final edited image.

Unlike methods relying on external segmentation, we utilize Token-wise Conditional Surprisal to quantify condition sensitivity based on the model's own counterfactual response. This metric measures the magnitude of prediction change under a condition switch; high values are treated as evidence that a token may require editing, whereas low values favor preservation. The score is a model-derived heuristic for editability rather than a semantic correctness certificate.

Specifically, we wish to decide, at each time $t$, which tokens should respond to the condition switch from $c_{\mathrm{src}}$ to $c_{\mathrm{tar}}$.

\paragraph{Same-state counterfactual evidence.}
We measure same-state conditional sensitivity with attention memory injection disabled. For coupled draw $\ell$, Eq.~\eqref{eq:z_src} uses $\boldsymbol{\varepsilon}^{(\ell)}$ to form $\mathbf{z}^{\mathrm{src},\ell}_t$, which is evaluated under both conditions:
\begin{equation}
    \hat{\mathbf{x}}_{0,\mathrm{src}}^{(\ell)} = \hat{\mathbf{x}}_0(\mathbf{z}^{\mathrm{src},\ell}_t,t,c_{\mathrm{src}}), \quad
    \hat{\mathbf{x}}_{0,\mathrm{tar}}^{(\ell)} = \hat{\mathbf{x}}_0(\mathbf{z}^{\mathrm{src},\ell}_t,t,c_{\mathrm{tar}}).
\end{equation}
Both predictions lie in $\mathbb{R}^{S\times d}$ and differ only in the condition for that draw.

For each token index $i\in\{1,\ldots,S\}$, let $\hat{\mathbf{x}}_{0,\mathrm{src}}^{(i,\ell)}$ and $\hat{\mathbf{x}}_{0,\mathrm{tar}}^{(i,\ell)}$ denote the predictions obtained from the $\ell$-th coupled noise draw. Noise is shared within each source/target pair and independent across $\ell$. We define Token-wise Conditional Surprisal by the Monte Carlo average
\begin{equation}
    s_i(t) \triangleq \frac{1}{n_{\mathrm{avg}}}\sum_{\ell=1}^{n_{\mathrm{avg}}}
    \frac{1}{d}\left\lVert \hat{\mathbf{x}}_{0,\mathrm{tar}}^{(i,\ell)} - \hat{\mathbf{x}}_{0,\mathrm{src}}^{(i,\ell)} \right\rVert_1,
    \label{eq:surprisal}
\end{equation}
where $\|\cdot\|_1$ is the $\ell_1$ norm in $\mathbb{R}^d$. The term surprisal is descriptive here and does not refer to Shannon surprisal. For each draw, Eq.~\eqref{eq:x0_from_v} gives
\begin{equation}
    \hat{\mathbf{x}}_{0,\mathrm{tar}}^{(i,\ell)} - \hat{\mathbf{x}}_{0,\mathrm{src}}^{(i,\ell)}
    = t\left( v_\theta(\mathbf{z}^{\mathrm{src},\ell}_t,t,c_{\mathrm{src}})^{(i)} - v_\theta(\mathbf{z}^{\mathrm{src},\ell}_t,t,c_{\mathrm{tar}})^{(i)} \right),
\end{equation}
so $s_i(t)$ averages the magnitude of the same-state velocity change induced by switching only the condition.

\paragraph{Top-quantile mask with smoothing.}
We select a sparse editable set by taking the $B(t)=\max\{1,\lfloor\rho(t)S\rfloor\}$ largest values of the EMA-smoothed scores $\{\bar{s}_i(t)\}_{i=1}^S$, where $\rho(t)\in(0,1]$ controls the editable fraction. Ties at the cutoff are resolved by a fixed token-index order, yielding exactly $B(t)$ indices and the mask $m_i(t)=1$ for $i\in\mathcal{E}(t)$ and $0$ otherwise. Before optional dilation, this top-$B(t)$ set maximizes $\sum_{i\in\mathcal{E}(t)}\bar{s}_i(t)$ among sets of the same cardinality; no stronger optimality claim is implied.

To reduce estimator noise across steps, we apply an exponential moving average (EMA) to the Monte Carlo score above. Let $\bar{s}_i(t)$ be the EMA buffer, initialized as $\bar{s}_i=0$. Upon receiving a new estimate $s_i(t)$, we update
\begin{equation}
    \bar{s}_i \leftarrow \beta\,\bar{s}_i + (1-\beta)\,s_i,
    \label{eq:ema}
\end{equation}
where $\beta\in[0,1)$ is the smoothing coefficient.

Mask updates are performed only within a mid-noise detect window $t\in[t_{\text{lo}},t_{\text{hi}}]$. To save compute, we update the mask every $\kappa$ steps (integer interval $\kappa\ge 1$). Optionally, we apply $r\ge 0$ rounds of morphological dilation on the 2D token grid to encourage spatial coherence; the keep set remains the complement after dilation.

\subsection{Attention Memory Boundary Condition via Partial Attention Memory Freezing}
\label{sec:mem-freezing}

Since standard diffusion models couple tokens globally through self-attention~\cite{vaswani2017attention}, simply freezing token states does not control every message-passing pathway. Replacing the keys and values of keep tokens makes those token positions supply source-branch memory in the target attention readout. This boundary does not freeze target queries or attention weights and does not remove global coupling; it limits one route by which target-conditioned information enters through keep-token memory.

A key challenge is enforcing that keep tokens preserve background and structure under the target condition $c_{\mathrm{tar}}$. We implement a boundary condition directly in the self-attention mechanism of the transformer.

\paragraph{Partial attention memory freezing (KV boundary).}
The memory-supply boundary is a concrete, token-aligned realization inspired by key/value reuse in prior attention-control methods~\cite{zhu2025kv}.

Consider one attention head with query and key matrices $Q,K\in\mathbb{R}^{S\times d_h}$ and value matrix $V\in\mathbb{R}^{S\times d_v}$.
During target-branch inference, we maintain two copies of attention memory:
$(K^{\mathrm{src}},V^{\mathrm{src}})$ computed from the source branch under $c_{\mathrm{src}}$, and
$(K^{\mathrm{tar}},V^{\mathrm{tar}})$ computed from the target branch under $c_{\mathrm{tar}}$.
Let ``keep'' denote token indices with $m_i(t)=0$ (equivalently $k_i(t)=1$).
We construct mixed memories $\widetilde K,\widetilde V$ by replacing target rows at keep-token positions:
\begin{equation}
(\widetilde K_j,\widetilde V_j)=
\begin{cases}
(K^{\mathrm{src}}_j,V^{\mathrm{src}}_j), & j\in\mathcal{K}(t),\\
(K^{\mathrm{tar}}_j,V^{\mathrm{tar}}_j), & j\in\mathcal{E}(t).
\end{cases}
\label{eq:mem_freeze}
\end{equation}
Queries remain untouched: the target branch uses $Q^{\mathrm{tar}}$ computed under $c_{\mathrm{tar}}$. Thus keep-token positions supply source value content, while target queries and the resulting attention weights remain condition dependent.

\subsection{Asynchronous Target-Conditioned Update}
\label{sec:update}

We now construct the update rule that follows target-conditioned dynamics only on editable tokens, evaluated under aligned randomness and under the attention boundary.

At each reverse step, let $\mathbf{m}(t)\in\{0,1\}^S$ be the editable mask vector at time $t$ and $\mathbf{k}(t)=\mathbf{1}-\mathbf{m}(t)$ the keep indicator.
We evaluate a source velocity at $\mathbf{z}^{\mathrm{src}}_t$ under $c_{\mathrm{src}}$, and a target velocity at the coupled $\mathbf{z}^{\mathrm{tar}}_t$ under $c_{\mathrm{tar}}$ with the KV boundary condition enabled for keep tokens.
For coupled draw $\ell$, define
$\Delta v_t^{(\ell)}=v_\theta(\mathbf{z}^{\mathrm{tar},\ell}_t,t,c_{\mathrm{tar}})-v_\theta(\mathbf{z}^{\mathrm{src},\ell}_t,t,c_{\mathrm{src}})$,
where the target evaluation uses the mixed memory in Eq.~\eqref{eq:mem_freeze}. The steering estimator is
\begin{equation}
    \widehat{\Delta v}_t \triangleq \frac{1}{n_{\mathrm{avg}}}
    \sum_{\ell=1}^{n_{\mathrm{avg}}}\Delta v_t^{(\ell)}
    \in \mathbb{R}^{S\times d}.
    \label{eq:delta_v}
\end{equation}
We update only editable tokens in the $x_0$-space state:
\begin{equation}
    \mathbf{x}^{\mathrm{edit}}_{t_{j+1}} \leftarrow
    \mathbf{x}^{\mathrm{edit}}_{t_j} + \Delta t_j \bigl( \widehat{\Delta v}_{t_j} \odot \mathbf{m}(t_j) \bigr).
    \label{eq:masked_update}
\end{equation}

Finally, we enforce the state-space contract by projection, implemented as a hard reset on keep tokens:
\begin{equation}
    \left(\mathbf{x}^{\mathrm{edit}}_{t_{j+1}}\right)_{\mathcal{K}(t_j)} \leftarrow \left(\mathbf{x}^{\mathrm{src}}_{t_j}\right)_{\mathcal{K}(t_j)}.
    \label{eq:hard_reset}
\end{equation}
Equation~\eqref{eq:hard_reset} is the exact Euclidean projection onto $\mathcal{C}(t_j)$ because the objective separates over keep and editable rows.
In practice, attention memory freezing \eqref{eq:mem_freeze} and projection \eqref{eq:hard_reset} are complementary:
the former constrains the memory supplied by keep-token positions but does not eliminate global message passing, while the latter enforces zero drift for the selected keep-token latent rows after each projection.

\subsection{Scope of Formal Guarantees}
\label{sec:formal_scope}

\paragraph{Parallel update scope.}
For a deterministic scheduler, the reverse trajectory has the form $\mathbf{z}_{t_{j+1}}=\mathcal{T}_{j,\theta,c}(\mathbf{z}_{t_j})$, so it is a Markov recursion on the full state $\mathbb{R}^{S\times d}$ and produces all token rows simultaneously. This establishes parallel token-state updates, not independence: $\mathcal{T}_{j,\theta,c}$ can couple every row through self-attention.

\paragraph{Common-noise scope.}
Let $F(\boldsymbol{\varepsilon})$ and $G(\boldsymbol{\varepsilon})$ be any scalar projections of the target and source evaluations with finite second moments. A one-sample common-noise difference and an independent-noise difference satisfy
\begin{align}
\mathrm{Var}[F(\boldsymbol{\varepsilon})-G(\boldsymbol{\varepsilon})]
&=\mathrm{Var}(F)+\mathrm{Var}(G)-2\mathrm{Cov}(F,G),\\
\mathrm{Var}[F(\boldsymbol{\varepsilon})-G(\widetilde{\boldsymbol{\varepsilon}})]
&=\mathrm{Var}(F)+\mathrm{Var}(G),
\end{align}
where $\widetilde{\boldsymbol{\varepsilon}}$ is independent. Thus common noise has no larger variance when $\mathrm{Cov}(F,G)\ge0$, and strictly smaller variance when the covariance is positive. Averaging $n_{\mathrm{avg}}$ independent coupled differences reduces their mean-squared fluctuation around the expectation by the standard factor $1/n_{\mathrm{avg}}$.

\paragraph{Attention-memory scope.}
For target query row $q_i^{\mathrm{tar}}$, define
$\widetilde{\alpha}_{ij}=\operatorname{softmax}_j((q_i^{\mathrm{tar}})^\top\widetilde K_j/\sqrt{d_h})$.
The output row $o_i=\sum_{j=1}^{S}\widetilde{\alpha}_{ij}\widetilde V_j$ then has the exact decomposition
\begin{equation}
o_i=\sum_{j\in\mathcal{K}(t)}\widetilde{\alpha}_{ij}V_j^{\mathrm{src}}+\sum_{j\in\mathcal{E}(t)}\widetilde{\alpha}_{ij}V_j^{\mathrm{tar}}.
\end{equation}
This identifies which values are supplied by keep-token positions, but $\widetilde{\alpha}_{ij}$ still depends on target queries and mixed keys; memory freezing therefore does not guarantee an unchanged attention output.

\paragraph{Selection and projection scope.}
For nonnegative scores $\bar{s}_i(t)$ and a fixed budget $B$, the pre-dilation top-$B$ set maximizes $\sum_{i\in\mathcal{E}}\bar{s}_i(t)$ among sets with $|\mathcal{E}|=B$: replacing any selected lower-score index by an unselected higher-score index cannot decrease the objective. This statement concerns captured score mass only. For projection, the Frobenius objective separates over $\mathcal{K}(t)$ and $\mathcal{E}(t)$; consequently, the unique minimizer fixes the keep block to $(\mathbf{x}^{\mathrm{src}}_t)_{\mathcal{K}(t)}$ and leaves the editable block equal to the pre-projection value. The resulting zero-drift statement is exact for the selected latent rows in exact arithmetic, but it does not by itself imply pixel-level invariance after decoding.

\section{Experiments}
\label{sec:experiments}

We evaluate ATDEdit on instruction-guided image editing. Since ATDEdit is an inference-time framework, all results are obtained without model fine-tuning.

\begin{table*}[t!]
\centering
\caption{\textbf{Quantitative comparison on PIE-Bench.} Methods are reported on their released/native backbones; FlowEdit, FlowAlign, and ATDEdit form the primary same-backbone SD3 comparison, while the remaining rows provide cross-backbone references. The best, second, and third numeric results are highlighted in green, yellow, and blue, respectively.}
\label{tab:main_comparison}
\resizebox{\textwidth}{!}{%
\begin{tabular}{@{}l l | c | cccc | cc@{}}
\toprule
\multirow{2}{*}{\textbf{Type}} & \multirow{2}{*}{\textbf{Method}} & \textbf{Struct.} & \multicolumn{4}{c|}{\textbf{Background Preservation}} & \multicolumn{2}{c}{\textbf{CLIP Semantics}} \\
\cmidrule(lr){3-3} \cmidrule(lr){4-7} \cmidrule(lr){8-9}
& & \textbf{Dist.} $(\times10^3)\downarrow$ & \textbf{PSNR}$\uparrow$ & \textbf{MSE} $(\times10^3)\downarrow$ & \textbf{SSIM} $(\times10^2)\uparrow$ & \textbf{LPIPS} $(\times10^3)\downarrow$ & \textbf{Whole}$\uparrow$ & \textbf{Edited}$\uparrow$ \\
\midrule
\multirow{6}{*}{Diffusion}
& DiffEdit (SD1.4) & 22.39 & 24.09 & 5.14 & 76.72 & 80.96 & 23.34 & 20.44 \\
& DDIM + MasaCtrl & 28.79 & 21.25 & 8.58 & 80.11 & 106.59 & 24.13 & 21.13 \\
& Direct Inversion + MasaCtrl & 24.46 & 21.78 & 7.99 & 81.74 & 87.38 & 24.42 & 21.38 \\
& DDIM + PnP & 28.20 & 21.26 & 8.42 & 78.90 & 113.58 & 25.45 & \cellcolor{RankSecond}22.54 \\
& Direct Inversion + PnP & 24.27 & 21.43 & 8.10 & 79.52 & 106.26 & 25.48 & \cellcolor{RankFirst}22.63 \\
& InfEdit (SD1.4) & 18.06 & 25.62 & 5.88 & 85.02 & \cellcolor{RankSecond}55.69 & 24.92 & 22.08 \\
\midrule
\multirow{5}{*}{Flow}
& FlowEdit (SD3) & 17.06 & 21.52 & 8.33 & 86.09 & 129.54 & \cellcolor{RankFirst}25.80 & \cellcolor{RankThird}22.42 \\
& FlowAlign (SD3) & \cellcolor{RankSecond}12.87 & \cellcolor{RankSecond}26.83 & 4.87 & \cellcolor{RankSecond}91.53 & \cellcolor{RankThird}57.87 & \cellcolor{RankSecond}25.77 & 22.31 \\
& KVEdit (FLUX) & \cellcolor{RankThird}13.84 & \cellcolor{RankThird}26.31 & \cellcolor{RankSecond}3.65 & \cellcolor{RankThird}91.19 & 60.04 & 25.72 & 22.11 \\
& Stable Flow (FLUX) & 15.46 & 24.33 & \cellcolor{RankThird}4.64 & 89.45 & 92.08 & \cellcolor{RankThird}25.73 & 22.26 \\
& \textbf{ATDEdit (SD3)} & \cellcolor{RankFirst}\textbf{11.10} & \cellcolor{RankFirst}\textbf{27.44} & \cellcolor{RankFirst}\textbf{2.70} & \cellcolor{RankFirst}\textbf{93.03} & \cellcolor{RankFirst}\textbf{55.35} & \cellcolor{RankFirst}\textbf{25.80} & \textbf{22.40} \\
\bottomrule
\end{tabular}%
}
\end{table*}

\begin{figure*}[t]
    \centering
    \includegraphics[width=1\linewidth]{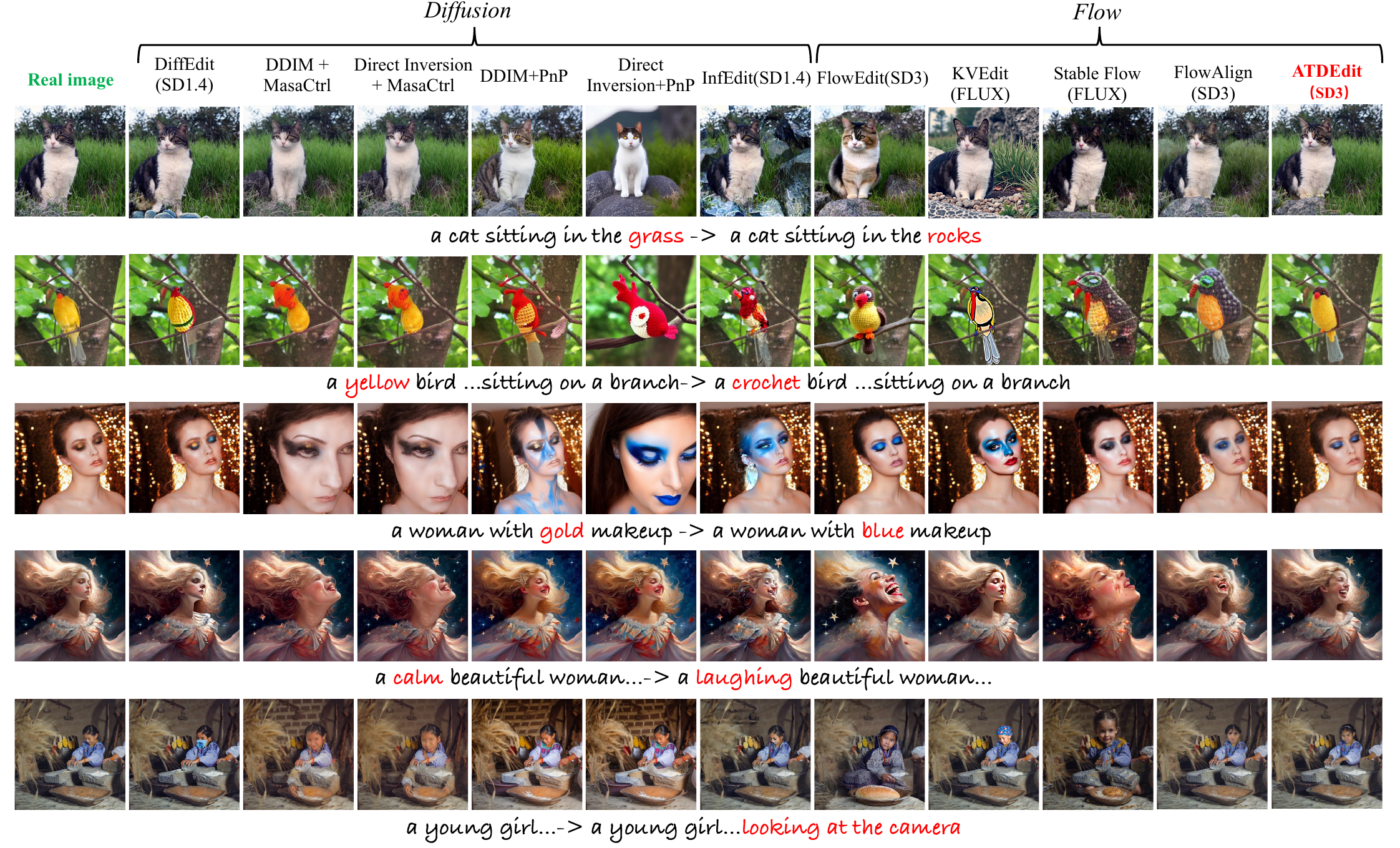}
    \caption{\textbf{Comparison of edited results.} Real images are in the first column.}
    \Description{A qualitative comparison grid. Each row begins with a real source image followed by results from multiple editing methods, allowing comparison of instruction alignment and preservation of unedited regions.}
    \label{fig:sota_grid}
\end{figure*}

\begin{figure}[!t]
    \centering
    \begin{subfigure}[b]{1\columnwidth}
        \centering
        \includegraphics[width=\linewidth]{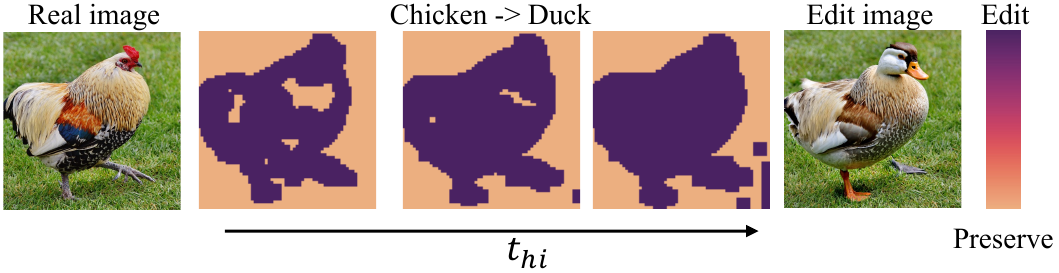}
        \caption{Editable-token masks at different $t_{\mathrm{hi}}$ values.}
        \label{fig:sub_heatmap}
    \end{subfigure}
    \vspace{-1mm}

    \begin{subfigure}[b]{0.49\columnwidth}
        \centering
        \includegraphics[width=\linewidth]{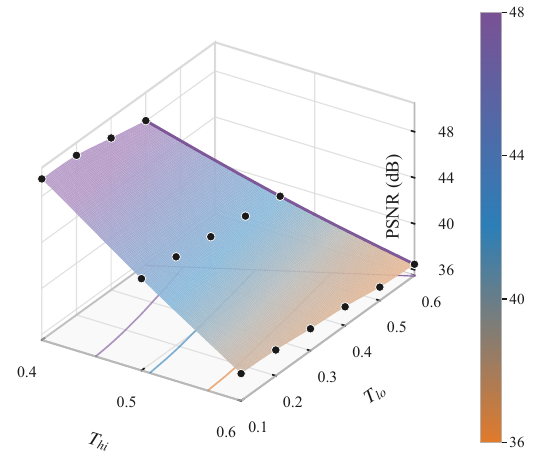}
        \caption{PSNR on unedited region $\uparrow$.}
        \label{fig:sub_psnr}
    \end{subfigure}%
    \begin{subfigure}[b]{0.49\columnwidth}
        \centering
        \includegraphics[width=\linewidth]{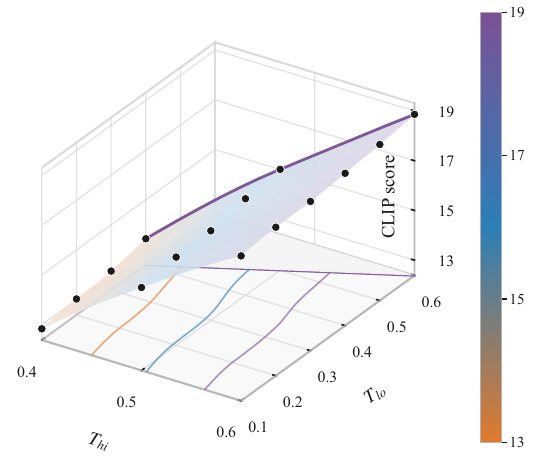}
        \caption{CLIP-Edited $\uparrow$.}
        \label{fig:sub_clip}
    \end{subfigure}
    \vspace{-1mm}
    \caption{\textbf{Sensitivity analysis.}
    (a) Binary editable-token masks obtained by top-quantile selection of the smoothed conditional sensitivity scores for the ``Chicken'' to ``Duck'' task.
    (b) PSNR on the unedited region over $(t_{\mathrm{hi}},t_{\mathrm{lo}})$.
    (c) CLIP-Edited over $(t_{\mathrm{hi}},t_{\mathrm{lo}})$.
    Samples occupy the valid triangular domain $t_{\mathrm{lo}}\leq t_{\mathrm{hi}}$; surfaces are interpolated only for visualization.
    The values in (b) and (c) are not dataset-level PIE-Bench averages.}
    \label{fig:ablation_sensitivity}
    \Description{Three sensitivity-analysis panels: binary editable-token masks for a chicken-to-duck edit, a three-dimensional surface of background PSNR over two thresholds, and a three-dimensional surface of edited-region CLIP score over the same thresholds.}
\end{figure}

\begin{table*}[!t]
\centering
\caption{\textbf{Ablation of the attention-memory boundary components.} Point estimates are reported; small numerical differences should be interpreted descriptively.}
\label{tab:main_ablation_boundary}
\resizebox{0.98\textwidth}{!}{%
\begin{tabular}{@{}l | cc | c | cccc | cc@{}}
\toprule
\multirow{2}{*}{\textbf{Variant}} & \multicolumn{2}{c|}{\textbf{Components}} & \textbf{Struct.} & \multicolumn{4}{c|}{\textbf{Background Preservation}} & \multicolumn{2}{c}{\textbf{Semantics}} \\
\cmidrule(lr){2-3} \cmidrule(lr){4-4} \cmidrule(lr){5-8} \cmidrule(lr){9-10}
 & \textbf{Mem. Freeze} & \textbf{Projection} & \textbf{Dist.} $(\times10^3)\downarrow$ & \textbf{PSNR}$\uparrow$ & \textbf{MSE} $(\times10^3)\downarrow$ & \textbf{SSIM} $(\times10^2)\uparrow$ & \textbf{LPIPS} $(\times10^3)\downarrow$ & \textbf{Whole}$\uparrow$ & \textbf{Edit}$\uparrow$ \\
\midrule
w/o Boundary & \xmark & \xmark & 19.46 & 21.41 & 8.17 & 78.47 & 110.13 & \cellcolor{RankFirst}25.91 & \cellcolor{RankFirst}22.53 \\
w/o Hard Proj. & \cmark & \xmark & \cellcolor{RankThird}13.20 & \cellcolor{RankThird}24.85 & \cellcolor{RankThird}5.11 & \cellcolor{RankThird}88.46 & \cellcolor{RankThird}72.37 & \cellcolor{RankSecond}25.86 & \cellcolor{RankSecond}22.47 \\
w/o Mem. Freeze & \xmark & \cmark & \cellcolor{RankSecond}11.52 & \cellcolor{RankSecond}27.27 & \cellcolor{RankSecond}2.83 & \cellcolor{RankSecond}92.48 & \cellcolor{RankSecond}60.12 & 25.46 & 21.90 \\
\textbf{ATDEdit (Full)} & \cmark & \cmark & \cellcolor{RankFirst}\textbf{11.10} & \cellcolor{RankFirst}\textbf{27.44} & \cellcolor{RankFirst}\textbf{2.70} & \cellcolor{RankFirst}\textbf{93.03} & \cellcolor{RankFirst}\textbf{55.35} & \cellcolor{RankThird}\textbf{25.80} & \cellcolor{RankThird}\textbf{22.40} \\
\bottomrule
\end{tabular}%
}
\end{table*}

\subsection{Experimental Setup}
\label{subsec:exp_setup}

\paragraph{Dataset and Evaluation Metrics.}
We evaluate on PIE-Bench~\cite{ju2023direct}, a standard benchmark for instruction-based image editing at $512\times512$ resolution. PIE-Bench contains 700 samples across 10 editing categories. Each instance provides a source image, an edit instruction, and a ground-truth region mask. ATDEdit does not use the ground-truth mask at inference time, and the mask is used only for evaluation.

We report three groups of metrics. Structure is measured by Structure Distance (Dist.) to assess geometric consistency~\cite{ju2023direct}. Background preservation is measured by PSNR, MSE, SSIM~\cite{wang2004image}, and LPIPS~\cite{zhang2018unreasonable} on the non-edited region. CLIP-Whole uses the full image, whereas CLIP-Edited keeps pixels inside the benchmark edited-region mask and zeros the complement before computing image--text similarity against the edit instruction~\cite{radford2021learning}. Dist., MSE, SSIM, and LPIPS are displayed after multiplication by the scale factors stated in the table headers.

\paragraph{Implementation Details.}
ATDEdit uses SD3~\cite{esser2024scaling} as the default backbone. For prior methods that are tightly coupled to specific released backbones, we follow their official implementations and report results on those native backbones. For SD3-based comparisons, we match the image resolution and the inference-step budget whenever possible. All experiments are conducted on a single RTX 3090.

\subsection{Comparison with Prior Methods}
\label{subsec:compare}

\paragraph{Baselines.}
We compare ATDEdit against ten representative baselines spanning several major paradigms of text-guided image editing. Specifically, the compared methods include diffusion-based editors such as DiffEdit~\cite{couairon2022diffedit} and InfEdit~\cite{xu2023inversion}, recent flow-based or accelerated editors including FlowEdit~\cite{kulikov2025flowedit}, FlowAlign~\cite{kim2025flowalign}, KVEdit~\cite{zhu2025kv}, and Stable Flow~\cite{avrahami2025stable}, as well as attention-control approaches based on MasaCtrl~\cite{cao2023masactrl} and Plug-and-Play (PnP)~\cite{tumanyan2023plug}. For fairness, we evaluate MasaCtrl and PnP under both DDIM~\cite{song2020denoising} and Direct Inversion~\cite{ju2023direct}, yielding four variants in total. Overall, these baselines cover a broad spectrum of editing paradigms, from multi-step diffusion editing and inversion-based reconstruction to recent flow-based generation and attention manipulation, providing a comprehensive benchmark across different model architectures and editing mechanisms.

\paragraph{Quantitative Results.}
Quantitative results on PIE-Bench are summarized in Table~\ref{tab:main_comparison}. In the reported comparison, ATDEdit obtains the strongest Structure Distance, PSNR, MSE, SSIM, and LPIPS, ties FlowEdit on CLIP-Whole, and remains competitive on CLIP-Edited. Because several prior methods use their native backbones, we treat FlowEdit and FlowAlign as the primary same-backbone SD3 comparisons and the other rows as cross-backbone references. The observed preservation gains are consistent with the intended roles of online token selection, source-memory replacement, and hard projection, but the table alone does not isolate backbone effects outside the SD3 subset.

\paragraph{Qualitative Results.}
Figure~\ref{fig:sota_grid} provides qualitative examples in which ATDEdit applies the requested edits while preserving much of the surrounding layout and background.

\subsection{Ablation Study}
\label{subsec:ablation}

\paragraph{Sensitivity analysis.}
We analyze the impact of the detection window $[t_{\text{lo}}, t_{\text{hi}}]$ on the trade-off between pixel-level fidelity and semantic alignment. The feasible parameter space is triangular because our design requires $t_{\text{lo}} \leq t_{\text{hi}}$.

Figure~\ref{fig:ablation_sensitivity} shows lower background PSNR but higher CLIP-Edited as either threshold increases. The stronger variation along $t_{\text{hi}}$ suggests that $t_{\text{hi}}$ is the primary knob for edit strength, whereas $t_{\text{lo}}$ mainly adjusts the baseline sensitivity.

\paragraph{Attention memory boundary condition.}
Table~\ref{tab:main_ablation_boundary} shows that memory freezing and hard projection are complementary rather than redundant.
Removing both components substantially weakens preservation, removing hard projection degrades every preservation metric, and removing memory freezing reduces both preservation and CLIP scores.
The measured effects are consistent with distinct roles: hard projection enforces the stated zero-drift constraint on selected latent rows, while memory freezing changes the source/target memory supplied during attention. The ablation does not directly measure boundary-local feature consistency.

\paragraph{Ablation takeaway.}
The sensitivity sweep characterizes the edit-strength trade-off, while the component-removal study supports distinct contributions from memory freezing and hard projection. These results do not establish higher-order synergy between components.

\section{Conclusion}
In this work, we present ATDEdit, an inference-time, token-level framework designed for image editing with Diffusion Transformers (DiTs). By conceptualizing the denoising process as a parallel yet globally coupled token-state update, our approach enables token-indexed condition switching to achieve fine-grained editing control. To effectively safeguard unedited regions, we implement source key/value memory substitution alongside hard projection on designated keep-token rows. Extensive evaluations on PIE-Bench demonstrate that ATDEdit attains the strongest reported background preservation metrics while maintaining competitive semantic alignment, and our boundary-component ablation studies further confirm the essential contributions of memory freezing and hard projection. However, ATDEdit incurs computational overhead during mask discovery and relies heavily on the DiT's calibration for accurate masking. Furthermore, its preservation-oriented hard projection is less suitable for global transformations unless most or all tokens are marked editable.

\section*{Acknowledgments}
This work was supported by Guangdong Basic and Applied Basic Research Foundation (Grant No. 2025A1515110445), National Training Program of Innovation and Entrepreneurship for Undergraduates (Grant No. 202511845040) and Guangdong University of Technology Training Program of Research for Undergraduates (Grant No. xj2026118450535).

\bibliographystyle{ACM-Reference-Format}
\bibliography{sample-base}

\end{document}